# AviationLMM: A Large Multimodal Foundation Model for Civil Aviation

Wenbin Li*
China Southern Digital Intelligence Technology (Guangdong) Co., Ltd.
China Southern Airlines Co., Ltd.
Guangzhou, China
li_wenbin@csair.com

Jingling Wu
China Southern Digital Intelligence Technology (Guangdong) Co., Ltd.
China Southern Airlines Co., Ltd.
Guangzhou, China
wujingl@csair.com

Xiaoyong Lin
China Southern Digital Intelligence Technology (Guangdong) Co., Ltd.
China Southern Airlines Co., Ltd.
Guangzhou, China
linxy@csair.com

Jing Chen
China Southern Digital Intelligence Technology (Guangdong) Co., Ltd.
China Southern Airlines Co., Ltd.
Guangzhou, China
chenjingc@csair.com

Cong Chen
China Southern Digital Intelligence Technology (Guangdong) Co., Ltd.
China Southern Airlines Co., Ltd.
Guangzhou, China
chenconga@csair.com

*Abstract*—Civil aviation is a cornerstone of global transportation and commerce, and ensuring its safety, efficiency and customer satisfaction is paramount. Yet conventional Artificial Intelligence (AI) solutions in aviation remain siloed and narrow, focusing on isolated tasks or single modalities. They struggle to integrate heterogeneous data such as voice communications, radar tracks, sensor streams and textual reports, which limits situational awareness, adaptability, and real‑time decision support. This paper introduces the vision of AviationLMM, a Large Multimodal foundation Model for civil aviation, designed to unify the heterogeneous data streams of civil aviation and enable understanding, reasoning, generation and agentic applications. We firstly identify the gaps between existing AI solutions and requirements. Secondly, we describe the model architecture that ingests multimodal inputs **such as** air‑ground voice, surveillance, on‑board telemetry, video and structured texts, **and** performs cross‑modal alignment and fusion, and produces flexible outputs ranging from situation summaries and risk alerts to predictive diagnostics and multimodal incident reconstructions. In order to fully realize this vision, we identify key research opportunities to address, including data acquisition, alignment and fusion, pretraining, reasoning, trustworthiness, privacy, robustness to missing modalities, and synthetic scenario generation. By articulating the design and challenges of AviationLMM, we aim to boost the civil aviation foundation model progress and catalyze coordinated research efforts toward an integrated, trustworthy and privacy‑preserving aviation AI ecosystem.

## I. Introduction

Over the last decade the civil aviation industry has embarked on an unprecedented wave of digitalization and intelligentization: flight management systems and autopilots have long assisted pilots, and AI approaches have emerged for tasks such as delay prediction, trajectory forecasting and anomaly detection [1]. These advances underpin a shift from reactive safety management toward proactive, data-driven strategies where AI can uncover patterns beyond human perception.

Despite progress, current AI applications in aviation remain narrow and fragmented. Conventional AI systems are largely task‑specific: separate models analyze flight sensor data for anomalies, transcribe pilot–controller radio conversations, or predict delays. Such siloed designs deliver useful insights but fail to provide a holistic situational picture or to cross-validate signals between modalities and tasks. Recent domain-specific language models, such as AviationGPT [2] and AviationLLM [3], have demonstrated the ability to improve aviation natural language tasks by fine-tuning large language models on curated aviation data. However, these systems remain text‑only, and cannot integrate non‑textual information such as operational videos, radar tracks or sensor telemetry. In practice, this leads to brittle performance and poor generalization across aircraft, airlines and operating conditions [4].

The last few years have ushered in a new era of large multimodal models capable of jointly processing and generating data across modalities. Models such as Flamingo [5], ImageBind [6] and LLaVA [7] link vision, language and other modalities through shared embeddings and instruction following; Next‑GPT extends this paradigm to any to any generation across text, image, video and audio modalities [8]. Remote‑sensing foundation models like SeaMo [9] and TerraMind [10] integrate spectral, spatial and meteorological inputs with season aware and generative architectures. Parameter efficient tuning frameworks such as AURORA [11] reduce the burden of adapting large models to new tasks by tuning just 0.04 % of model parameters, while personalized multimodal prompts such as VIP5 [12] enable cross domain recommendation with minimal training. These advances reveal the potential for task agnostic models that learn universal patterns and can generalize to new problems with few examples. This is a promising direction for aviation, where data modalities include speech, time-series telemetry, vision and structured text.

While the incorporation of multimodality presents a promising frontier, it also brings to light significant challenges that have yet to be fully addressed in aviation. Firstly, multimodal data from avionics, communication and operations systems have disparate sampling rates, noise characteristics and semantic structures. Aligning and fusing them requires sophisticated synchronization and representation learning [13]. Secondly, aviation data are scarce for rare events such as incidents or failures, and the high cost of labeling limits supervised learning. Thirdly, safety, trust and reliability are paramount: recent evaluations of foundation models highlight vulnerabilities such as hallucinations, bias and privacy leaks [14][15]. Conformal alignment frameworks show how controlling false discovery rates can certify outputs in high-stakes domains [16]. Fourthly, large models demand substantial computational resources, making real-time inference and deployment challenging. Parameter-efficient tuning and edge–cloud splits help but require further research. Furthermore, many multimodal models assume all modalities are present; when some inputs are missing due to sensor failures or privacy constraints, performance can degrade significantly [17], prompting research on correlated prompts for missing modalities [18]. Finally, data governance and regulatory certification necessitate privacy-preserving training and auditable decision traces.

In this work, we propose the idea and the vision of AviationLMM (Large Multimodal foundation Model for civil Aviation), a foundation model to transform civil aviation analytics by integrating multi-modal inputs not only to synthesize a coherent operational picture in real time but also to deliver comprehensive multi-task analyses with a variety of output modalities. The concept extends contemporary large language model (LLM) and large multimodal model (LMM) advances with task-oriented multimodal prompting coupled with human/AI feedback, a hybrid training paradigm that leverages labeled, unlabeled and weakly labeled data to improve generalization under rarity and shift, and an edge–cloud split that runs privacy-sensitive encoders close to data sources while fusing and decoding in secure regional clouds. Together with the encode–align–fuse–decode model structure, these ingredients aim to unify speech, surveillance, telemetry, video and operational text into a single, queryable state that supports situation awareness, risk monitoring, predictive diagnostics and service interactions.

The remainder of the paper is organized as follows: Section 2 reviews the gaps between the civil aviation requirements and existing AI solutions; section 3 describes the AviationLMM concept highlighting its capability across modalities and tasks; section 4 presents the corresponding research opportunities to fully achieve the AviationLMM's potential; section 5 sketches potential application scenarios, and section 6 concludes the paper.

## II. GAPS IN CURRENT AI SOLUTIONS FOR CIVIL AVIATION

### A. Fragmented and uni-modal solutions

Existing aviation AI systems are still built as single-modality, single-task pipelines, which prevents holistic situational awareness. Along the operations stack, anomaly detection commonly targets flight-data-recorder (FDR) channels in isolation, without explicit ties to surveillance or communications streams [19]. In parallel, Automatic Speech Recognition (ASR) pipelines focus on transcribing Air Traffic Control (ATC) to pilot radio exchanges and intent parsing but generally remain confined to the speech modality [20]. On the scheduling side, delay predictors largely rely on structured variables (e.g., timetable and weather) to engineer features and build models [21]. While these siloed systems are effective within their niches, they lack cross modal cross checks and cannot aggregate causal signals across channels. To push beyond purely textual capabilities, domain-specific LLMs such as AviationGPT [2] and AviationLLM [3] finetune on curated aviation corpora and demonstrably improve document Q&A and training dialogues; nevertheless, they remain text-only and do not ground answers in sensor, track, or video evidence, leaving multi-source situational awareness unaddressed. A unified multimodal representation and reasoning framework is needed, to perform alignment and fusion across speech, tracks, telemetry, video, and text within a single model, enabling consistent situation pictures and mutual corroboration of evidence across tasks.

### B. Limited generalization and data scarcity

Conventional models are often developed to a specific airline or airspace scenario and face data scarcity for rare events, which limits supervised learning and out-of-distribution robustness. Domain LLMs (e.g., AviationGPT [2], AviationLLM [3]) underscore their value on text tasks while simultaneously exposing the limited labeled resources and coverage gaps endemic to aviation. Within the speech sub-domain, the ATCO2 corpus contributes thousands of hours of ATC audio, strengthening acoustics and language modeling; yet, it still lacks the cross-modal, event-level annotations required to couple communications with surveillance or telemetry for joint learning [22]. Cross domain evidence suggests that curating multi-season, multi-source datasets, as seen in remote sensing, helps counter distribution shift and improve generalization [9]. Meanwhile, parameter-efficient tuning, such as AURORA [11], markedly reduces the cost of adaptation but does not by itself solve the sample sparsity for rare multimodal scenarios. Thus, we need hybrid training regimes combing supervised and unsupervised paradigms, simulation-based augmentation, together with standardized schemas and alignment labels, to lift generalization and stability across fleets, operators, and operating conditions.

### C. Trust, safety and reliability

As a high-assurance domain, aviation requires models whose outputs are explainable, robust, and certifiable, while generic

LMM vulnerabilities are amplified in this setting. Multimodal benchmarks such as HEMM [23], MMDT [14], and MultiTrust [15] consistently reveal hallucinations, factual mistakes, bias, and privacy leakage in advanced foundation models. In video understanding, Q-Bench [24] highlight systematic deficits in spatiotemporal reasoning and low-level visual fidelity. To provide statistical coverage and calibration, conformal alignment [16] demonstrates controlling false discovery rates as a path to certifiable outputs in high-stakes deployments. Yet, in aviation today, end-to-end systems that integrate uncertainty quantification, evidence tracing, and auditable explanations remain rare. Aviation-grade deployment demands domain-specific trustworthy evaluation, evidence-linked outputs, and calibrated uncertainty baked into the model stack and lifecycle.

*D. Handling missing modalities*

Real world operations frequently suffer missing or degraded modality inputs (e.g., sensor failure, bandwidth or privacy constraints, partial communications), while many multimodal models assume complete inputs. Surveys document that absent modalities significantly erode current models' performance and stability [17]. To mitigate this, deep correlated prompting [18] exploits cross modal and cross prompt correlations at inference time to compensate for missing inputs and reduce degradation. In practice, aviation systems often resort to dropping modalities or relying on hardware redundancy approaches that fall short for complex, interdependent scenarios. Training and inference must become modality-aware, using modality-dropout, reliability gating, and uncertainty propagation together with contextual reconstruction so the system degrades gracefully and remains safe when key inputs are unavailable.

*E. Resource demands and deployment challenges*

The compute and latency profile of large models clashes with aviation's real-time, bandwidth-limited, and edge-cloud constrained environments. On the adaptation front, parameter-efficient methods such as AURORA [11] reduce trainable parameters and memory; in adjacent multimodal tasks, lightweight adapters like VIP5 [12] achieve cross-domain transfer with minimal code changes. However, aviation additionally requires carefully engineered edge cloud splits, running privacy-sensitive encoders on edge and fusion and decoding in secure regional clouds. While split learning and federated learning offers tools for privacy and bandwidth governance [25-27], scaling them to high-risk, low-latency, multimodal aviation use cases still lack mature practice and systematic evaluation. We need architectures expressly designed for edge cloud collaboration with splittable encode-align-fuse-decode pipelines, comm-aware representation compression, and end-to-end latency budgets for compliant, scalable deployment.

*F. Absence of an integrated aviation foundation model*

No domain tailored multimodal foundation model exists to jointly learn from air ground voice, surveillance tracks, onboard telemetry, video feeds, and operational texts and supports any-to-any I/O for aviation. Existing efforts are primarily text-centric and excel within corpus boundaries but cannot yet integrate multi-sensor and visual streams. Generic multimodal approaches (e.g., Next-GPT [8], ImageBind [6]) provide any to any mappings and a shared embedding space, but lack aviation knowledge and operational constraints. By comparison, in other domains, AURA-MFM [28] shows that incorporating additional sensor modalities (e.g., IMU, motion capture) and aligning them with text and video improves complex activity understanding, indirectly evidencing the value of domain-specific multimodal integration. This filed needs an aviation-aware multimodal foundation model, embedding domain knowledge and procedural and physical constraints that perform cross-modal alignment and fusion in a unified representation and realizes any to any, auditable outputs across safety, operations, and service scenarios.

## III. AviationLMM Overview

AviationLMM is conceived as a scalable, task agnostic platform capable of end-to-end reasoning over heterogeneous aviation data. Its core follows an encode–align–fuse–decode pipeline [29] as depicted in fig. 1: (i) raw data are preprocessed and encoded into modality-specific latent spaces; (ii) these representations are temporally, spatially, and semantically aligned; (iii) aligned features are fused into a unified operational state; and (iv) multiple decoders generate outputs tailored to diverse downstream tasks. This pipeline operates in an edge cloud collaboration schema to respect privacy, latency, and bandwidth constraints: sensors and initial encoders run at the edge (e.g., onboard aircraft, at towers, in airline operation centers), while heavier alignment, fusion, and decoding modules run in secure cloud regions.

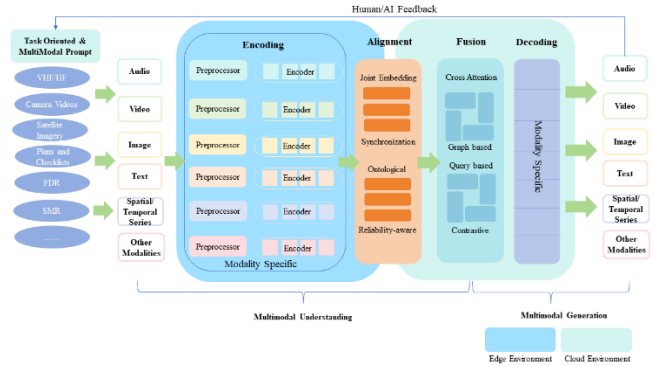

Fig.1. AviationLMM overview

- **Inputs.** The model handles six modality classes: audio covers Air Traffic Control (ATC) and pilot communications on Very High Frequency/High Frequency (VHF/HF), cabin announcements, and engine/environment sounds. Video includes tower and apron feeds, cockpit and cabin cameras, and digital-tower sensors. Image encompasses single frames from cameras, satellite imagery, weather composites and maintenance borescope and thermal imagery. Text comprises Notices to Air Missions (NOTAMs), Meteorological Aerodrome Reports (METARs), Terminal Aerodrome Forecasts (TAFs), flight plans, checklists, Maintenance/Configuration Deviation Lists (MEL/CDLs), ACARS/CPDLC messages and reports. Spatial/temporal series consist of Flight Data Recorder (FDR) and Health and Usage Monitoring System (HUMS) channels, Automatic Dependent Surveillance–

Broadcast (ADS-B) and radar tracks, telematics and numerical weather predictions. Other modalities cover volumetric weather radar, Light Detection and Ranging (LiDAR) or Surface Movement Radar (SMR) for ground movements, inertial measurement units (IMUs), and environmental sensors.

- **Preprocessing.** Since inputs can be asynchronous or partially missing among modalities, AviationLMM aligns and processes them dynamically. Prior to encoding, each modality undergoes tailored cleansing and anonymization. Audio is denoised, diarized and segmented; video and image streams are stabilized, resized and privacy - filtered; text is tokenized and augmented with aviation lexicons; sensor data are resampled and filtered, with synchronized time stamps and geo - coordinates. These steps harmonize heterogeneous inputs and provide data essential for subsequent alignment.

- **Encoders.** AviationLMM applies specialized encoders to each modality: audio encoders draw on WavLM-style self-supervised models that capture context beyond phonemes [30]; video encoders employ vision-transformer backbones with masked-video pretraining to learn spatio-temporal features [13]; image encoders rely on self-distillation architectures such as DINOv2 to capture visual semantics [31]; text encoders adopt instruction-tuned large language models that understand aviation jargon and procedures [32]; spatial/temporal series encoders use transformer-based models to capture long-range dependencies in FDR/HUMS signals, radar tracks and telematics, often integrating IMU data into a unified temporal representation [33]; and other-modality encoders apply 3D convolutional networks and positional encoding to volumetric weather radar and LiDAR/SMR scenes, enabling consistent latent representations [34].

- **Alignment.** To co-register heterogeneous embeddings, AviationLMM relies on state-of-the-art alignment techniques. Temporal–spatial synchronization with cross-modal attention learns to match asynchronous events across time and space [35]. Joint embedding models, like ImageBind, project different modalities into a shared space for unified retrieval and matching [6]. Ontological alignment can introduce domain ontologies (e.g., airspace, aircraft systems) to map semantically related cues; reliability-aware alignment assigns confidences to each modality based on signal quality and missingness. These methods allow the system to associate voice instructions with trajectory changes and sensor anomalies, despite differing sampling rates and noise profiles.

- **Fusion.** After alignment, AviationLMM fuses multimodal embeddings into a unified operational state following the dominant approaches: cross attention-based transformers enable multi-head attention across modalities, capturing dependencies over long horizons [13]. Graph based models represent entities (aircraft, runways, ground vehicles) as nodes, with modality-specific features on edges; this structure supports relational reasoning and safety rule enforcement. Hierarchical fusion builds layered representations, first integrating sensor and track data into state vectors, then blending text and visual context. Contrastive fusion strategies use cross-modal contrastive losses to bind modalities and maintain representation consistency [28]. Furthermore, the query-based fusion [36] introduced a mixed query strategy that unifies learnable and conditional queries enabling a single architecture to handle multiple segmentation tasks and datasets with improved generalization. These strategies ensure salient information is preserved while enabling robust cross-modal reasoning.

- **Decoders.** To support any-to-any outputs, AviationLMM includes modality-specific decoders: Audio decoders generate spoken advisories or re-enacted communications, leveraging advanced speech synthesis models [37]. Video decoders synthesize synchronized video reconstructions or animations, adapting recent video-generation techniques to multimodal contexts [38]. Image decoders produce anomaly heatmaps, trajectory overlays or inspection visualizations, often using diffusion models conditioned on fused latent embeddings [39]. Text decoders (i.e., adapted LLMs) output narrative reports, procedural steps and Q&A responses tailored to aviation tasks. Spatial/temporal series decoders create numerical outputs such as risk scores, remaining-useful-life curves, and trajectory forecasts via calibrated regression and classification heads [40]. Other modality decoders reconstruct volumetric weather or LiDAR/SMR data and generate synthetic sensor signals.

- **Edge Cloud Collaboration.** To respect privacy, latency and bandwidth constraints, AviationLMM partitions computation. Encoders run on aircraft avionics, tower servers and airline facilities, producing compressed latent representations. While alignment can be run either on edge or cloud environment according to the specific configurations, fusion and decoding occur in regional clouds. Split learning allows collaborative training without exposing raw data [25], and federated foundation-model methods enable cross organization learning with secure aggregation and differential privacy [27]. This configuration supports incremental updates and resilient operation even under intermittent connectivity.

- **Task-Oriented and Multi-Modal Prompting.** AviationLMM incorporates task-oriented prompts that combine textual instructions with multimodal inputs (e.g., voice snippets, trajectory segments, images) to specify desired outputs. Inspired by instruction-tuned models [7] and the idea of prompts with feedback [41][42], prompts can be refined via reinforcement learning from human or AI feedback to optimize task performance and safety. This mechanism provides high-level control without retraining the entire model, enabling flexible task composition while allowing dynamic customization with results grounded in multimodal evidence.

- **Hybrid Pretraining.** AviationLMM's training recipe combines supervised, self-supervised and weakly supervised objectives: labeled incident and maintenance datasets provide supervision; masked-signal and autoencoding objectives leverage unlabeled logs [43]; contrastive and weakly supervised alignment exploit weak labels; synthetic data from simulators and generative models augment rare events. This hybrid approach allows AviationLMM to learn from scarce labeled data while incorporating vast unlabeled and synthetic corpora.

## IV. RESEARCH OPPORTUNITIES

To fully achieve the vision of AviationLMM, the following research opportunities address key challenges beyond architecture design.

### A. Opportunity 1: Advanced alignment and fusion mechanisms

The key research opportunity for LMM is to deliver reliability-aware alignment and fusion that reconcile asynchronous clocks, heterogeneous uncertainty, and domain semantics (e.g., procedures, systems, airspace) into a unified, long-context operational state that preserves causality and remains queryable and certifiable. Although surveys formalize the open challenges of multimodal alignment or fusion [13] and models with dedicated sensor branches show the value of careful, contrastive fusion, future work should focus on confidence-weighted alignment with uncertainty propagation and latency compensation, graph-based cross-modal attention that associates events across space–time, ontology-guided process grounded in aircraft systems and checklists, and explicit causal-consistency checks that prevent spurious cross-modal shortcuts.

### B. Opportunity 2: Multimodal data collection, standardization, and simulation

Building a living aviation data fabric is highly desired that continuously captures, synchronizes, and anonymizes voice, surveillance, telemetry, video, and operational text across fleets, airports, seasons, and weather regimes, with an emphasis on event-level co-registration, rare-event coverage, and schema consistency so downstream alignment, reasoning, and evaluation can be automated rather than handcrafted. While speech-centric corpora ATCO2 demonstrate scalable pipelines but remain modality-limited [22], and remote-sensing efforts SeaMo [9], TerraMind [10] suggest that multi-season, multimodal curation and any-to-any generation improve robustness, the key next step is to establish a public–private alliance that defines harmonized schemas and event ontologies, codifies time–space alignment protocols and quality metrics, and operationalizes privacy oriented synthetic data toolchains (e.g., scenario graphs, dynamics simulators) with domain expert review and strong dataset governance.

### C. Opportunity 3: Efficient pretraining and domain adaptation

Another target to achieve is the foundation-scale capability under aviation's data scarcity and compute limits, enabling frequent re-alignment to fleets, routes, seasons, and procedures without full retraining. While hybrid objectives can scale learning on weak labels [44], and parameter-efficient tuning AURORA [11] and lightweight adapters VIP5 [12] reduce adaptation cost, with synthetic generators expanding the long tail, the path forward is to devise curricula that interleave real and synthetic data to emphasize rare event competence, standardize finetuning for operator variants with frozen backbones, craft aviation specific self-supervised pretexts (e.g., trajectory voice alignment, procedural step prediction), and develop data filtering strategies that control scale while retaining edge cases.

### D. Opportunity 4: Multimodal reasoning and knowledge integration

The opportunity is to move from correlation to causal, procedure aware reasoning over synchronized evidence, explaining why a risk is rising, which rule or envelope is implicated, and how to mitigate, and then verifying the chain with simulators or digital twins. Although holistic MLLM evaluations [23] reveal mixed performance on reasoning and knowledge grounding, and surveys [45] motivate tool use, neuro-symbolic links, and stepwise decomposition, progress now depends on hybrid neuro-symbolic controllers that encode flight-dynamics envelopes, and scenario specific rules, tool augmented inference that calls dispatch tools, trajectory solvers, and traffic simulators in a verify and validate loop, and transparent intermediate artifacts (e.g., timelines, causal graphs) that support investigation, training, and regulatory review.

### E. Opportunity 5: Trustworthiness, safety, and ethics

This opportunity is to establish certification grade trust pipelines that calibrate uncertainty, trace outputs to evidence, quantify risk and utility under time pressure, and guarantee privacy and fairness across operators and regions. Current benchmarks expose hallucination, robustness, fairness, and privacy weaknesses in MLLMs and video LMMs [23][27], and conformal alignment [9] offers a way to control false discoveries with statistical guarantees. Research should create aviation-specific risk-aware metrics (e.g., time critical false alarms and cascading risk cost), build evidence-linked answers with provenance to data snippets, integrate calibrated uncertainty and reject options for human escalation, and then finally embed feedback loops and bias audits throughout the model lifecycle.

### F. Opportunity 6: Privacy preserving learning and federated deployment

The opportunity is to enable multi-party collaboration (e.g., airlines, air navigation service provider, equipment manufacturer) without raw-data sharing, while meeting jurisdictional and contractual constraints and edge cloud bandwidth limits. Although federated foundation models [26][27] outline aggregation, heterogeneity, and trust challenges and early solutions, and split learning [25] keeps raw data on-prem with server-side continuation, the next phase requires co-designed encode–align–fuse–decode splits with communication aware embedding compression, secure aggregation and differential privacy tailored to multimodal sequences, and formal exploration of the privacy, latency, accuracy trade space for operational deployments.

*G. Opportunity 7: Robustness to missing modalities*

The opportunity is to guarantee graceful degradation when sensors or links fail, privacy redactions apply, or inputs are delayed. Under such situation, maintaining situational awareness and safety margins without brittle collapse is critical. Surveys document large performance drops under absent modalities [17], and deep correlated prompting adapts pretrained models via prompt-level correlations [18]. Future research should integrate modality dropout into pretraining, apply reliability gating and uncertainty propagation at inference to re-weight evidence, and develop contextual reconstruction from available streams coupled with input-health monitoring and policy-driven fallbacks (e.g., speech and tracks only) with explicit performance guarantees.

*H. Opportunity 8: Synthetic data and scenario generation*

The opportunity is to use controllable generators to inject rare hazards, weather extremes, and traffic pathologies, and then to quantify how well models detect, reason, and recover under various scenarios. Although TerraMind [10] showcases any-to-any multimodal generation and thinking in modalities in Earth observation, aviation needs domain-specific generators and digital twins that co-synthesize radar, comms, and video under weather, procedure and failure controls, with validation and coverage metrics (diversity, difficulty, plausibility). The aim is to integrate these generators into training pipelines, stress-testing harnesses, and safety cases.

## V. Application Scenarios

Upon steady progress on the research opportunities above and a fully AviationLMM, a broad set of multimodal, any-to-any application scenarios will become feasible across safety, operations, maintenance, airport management, and customer service. Below we present five representative application scenarios. For each, we open with a concise description and then summarize the design in a table with four fields: Inputs & Modality, Outputs & Modality, Primary Users, and Runtime & Deployment.

*A. Air Traffic Control Sector Assistant*

Air traffic control sector assistant uses radio communications, surveillance tracks, weather, and optional tower video to maintain a live sector picture, surfaces trajectory deviations, miscommunications, and sequencing risks, and proposes deconfliction actions with uncertainty and linked evidence. The characteristics of this scenario are described in Table I.

TABLE I: Scenario Characteristics Of Air Traffic Control Sector Assistant

| Inputs & Modality | Outputs & Modality | Primary Users | Runtime & Deployment |
|---|---|---|---|
| -audio: Air Traffic Control (ATC), pilot Very High Frequency/High Frequency (VHF/HF) -spatial/temporal series: Automatic Dependent Surveillance–Broadcast (ADS-B), primary/secondary radar tracks -text: Notices to Air Missions (NOTAM), Meteorological Aerodrome Reports (METAR), Terminal Aerodrome Forecasts (TAF) -video: tower camera | -text: sector summaries, conflict lists -audio: Text-to-Speech (TTS) advisories -image: timeline, map overlays | -ATC controllers; -sector supervisors | sub-second to seconds; edge encoders plus regional cloud fusion |

*B. Flight Deck Decision Support*

Flight deck decision support acts as a virtual co-pilot that correlates FDR HUMS signals, flight plans, and crew ATC speech to anticipate and contextualize risk, and also provides procedure aware advisories. The characteristics of this scenario are described in Table II.

TABLE II: Scenario Characteristics Of Flight Deck Decision Support

| Inputs & Modality | Outputs & Modality | Primary Users | Runtime & Deployment |
|---|---|---|---|
| -spatial/temporal series: Flight Data Recorder (FDR), Health and Usage Monitoring System (HUMS) -audio: flight-crew and ATC communications -text: Standard Operating Procedures (SOPs), checklists -image: flight-deck panel, Engine Indicating and Crew Alerting System (EICAS) snapshot | -audio: spoken advisories -text: procedure steps, explanations -spatial/temporal series: trend dashboards, short-term forecasts | -Flight crew | 100–500 ms alerts; on-aircraft edge compute plus on-premises gateway |

*C. Predictive Maintenance*

Predictive maintenance learns fleet wide degradation signatures from sensor streams, work logs, borescope imagery, and engine acoustics, and then prioritizes interventions with RUL forecasts, fault attribution, and suggested parts. The characteristics of this scenario are described in Table III.

TABLE III: Scenario Characteristics Of Predictive Maintenance

| Inputs & Modality | Outputs & Modality | Primary Users | Runtime & Deployment |
|---|---|---|---|
| -spatial/temporal series: engine/airframe sensors -text: Minimum Equipment List (MEL), Configuration Deviation List (CDL), maintenance logs -image: borescope, thermal imagery -audio: engine acoustics | -text: work cards, fault attributions and explanations -spatial/temporal series: Remaining Useful Life (RUL) curves -image: anomaly, heatmaps | -Maintenance, Repair and Overhaul (MRO) engineers; -Maintenance Control Center (MCC) | batch to near-real-time; secure cloud plus airline data lake |

## D. Safety Incident Reconstruction

Safety incident reconstruction automatically reconstructs events by synchronizing CVR and ATC audio, FDR, ADS-B and radar series, videos, and reports into a time-aligned narrative, and then generates counterfactuals and training variants for rare-event rehearsal. The characteristics of this scenario are described in Table IV.

TABLE IV: Scenario Characteristics Of Safety Incident Reconstruction

| Inputs & Modality | Outputs & Modality | Primary Users | Runtime & Deployment |
|---|---|---|---|
| -spatial/temporal series: FDR, ADS-B, radar logs<br>-audio: Cockpit Voice Recorder (CVR), ATC communications<br>-video: tower/cabin cameras<br>-text: occurrence/incident reports | -video: synchronized replay<br>-text: narratives, checklist mappings<br>-image: trajectory/impact overlays<br>-audio: re-enacted communications | -safety investigators;<br>-training organizations | offline to near-real-time; secure sandbox plus cloud rendering |

## E. Airport Surface Operation Manager

Airport surface operation manager coordinates apron movements by fusing A-SMGCS ground radar, belt telematics, gate cameras, schedules, and surface sensors, and then recommends pushback or taxi sequencing and resource allocation to compress turnaround while minimizing incursions. The characteristics of this scenario are described in Table V.

TABLE V: Scenario Characteristics Of Airport Surface Operation Manager

| Inputs & Modality | Outputs & Modality | Primary Users | Runtime & Deployment |
|---|---|---|---|
| -video: apron/gate cameras<br>-spatial/temporal series: Advanced Surface Movement Guidance and Control System (A-SMGCS) ground radar, vehicle telematics<br>-text: Target Off-Block Time (TOBT), Target Start-Up Approval Time (TSAT), flight schedules<br>-others: Light Detection and Ranging (LiDAR), Surface Movement Radar (SMR) | -text: turn plans, sequencing recommendations<br>-audio: ramp advisories<br>-image: Gantt charts, surface-map overlays<br>-spatial/temporal series: Estimated Time of Arrival (ETA) predictions | -Airline Operations Control Center (AOCC);<br>-ramp control;<br>-ground operations | 1–10 s loop; edge video analytics plus cloud fusion/optimization |

## VI. Conclusion

This work has introduced AviationLMM, a large multimodal foundation model tailored to the complex and safety-critical domain of civil aviation. By unifying heterogeneous data streams within an encode–align–fuse–decode architecture, AviationLMM overcomes the fragmentation of traditional AI systems by establishing a shared representational framework that enables perception, reasoning, and generation across modalities, supporting any-to-any intelligence in air traffic control, flight operations, predictive maintenance, airport management, and service applications.

Through its technical design and conceptual framework, AviationLMM bridges several long-standing challenges in aviation AI: it advances multimodal alignment and fusion, enables scalable learning under data scarcity through hybrid and parameter-efficient pretraining, and embeds trustworthy and privacy-preserving mechanisms across the model lifecycle. The paper also articulated eight key research opportunities, from alignment and simulation data generation to trustworthiness and federated deployment, that together define a roadmap for transforming aviation analytics from fragmented solutions into a coherent, certifiable system of intelligence. Each opportunity calls for deeper collaboration between AI researchers, aviation engineers, regulators, and industry practitioners to ensure the safe and effective integration of large models into real-world aviation workflows.

Looking ahead, realizing the full potential of AviationLMM will require standardized benchmarks, privacy-compliant collaboration frameworks, and real-world edge–cloud deployments that validate performance under operational constraints. Integrating digital twins, neurosymbolic reasoning, and causal verification mechanisms will further enhance interpretability, reliability, and regulatory acceptance. Ultimately, AviationLMM represents a foundational step toward trustworthy, multimodal, and mission-critical AI in civil aviation—linking perception, understanding, and action to enable proactive safety management, intelligent operations, and human-centered decision support across the global air transport system.